\documentclass[10pt,twocolumn,letterpaper]{article}

\usepackage{iccv}     
\usepackage[accsupp]{axessibility}  

\definecolor{iccvblue}{rgb}{0.21,0.49,0.74}
\usepackage[pagebackref,breaklinks,colorlinks,allcolors=iccvblue]{hyperref}
\usepackage{booktabs} 
\usepackage{ragged2e}
\usepackage{tabularx}
\usepackage[ruled,vlined]{algorithm2e}
\usepackage{graphicx}
\usepackage{subcaption}
\usepackage{xcolor}

\usepackage{tabularx} 
\def\paperID{*****} 
\def\confName{ICCV}
\def\confYear{2025}

\title{Prompt-Guided Spatial Understanding with RGB-D Transformers for Fine-Grained Object Relation Reasoning}

\author{
Tanner W. Muturi$^{1}$\thanks{Equal contribution} \quad
Blessing A. Kyem$^{2}$\footnotemark[1] \quad
Joshua K. Asamoah$^{2}$\footnotemark[1] \quad
Neema J. Owor$^{1}$ \quad
Richard Dyzinela$^{3}$ \\
Andrews Danyo$^{2}$ \quad
Yaw Adu-Gyamfi$^{1}$ \quad
Armstrong Aboah$^{2}$\thanks{Corresponding author} \\
\vspace{0.5em}
$^1$University of Missouri–Columbia  \quad
$^2$North Dakota State University \quad
$^3$Texas A\&M\\
\vspace{0.3em}
\parbox{\textwidth}{\centering\footnotesize%
\texttt{\{nodyv, twmtyg, adugyamfiy\}@missouri.edu}\\
\texttt{\{blessing.kyem,joshua.asamoah,andrews.danyo,armstrong.aboah\}@ndsu.edu}\\
}\\
\vspace{0.3em}
\small{$^*$First Authors} \quad
\small{$^\dagger$Corresponding author}
}

\begin{document}
\maketitle
\begin{abstract}
Spatial reasoning in large-scale 3D environments such as warehouses remains a significant challenge for vision-language systems due to scene clutter, occlusions, and the need for precise spatial understanding. Existing models often struggle with generalization in such settings, as they rely heavily on local appearance and lack explicit spatial grounding. In this work, we introduce a dedicated spatial reasoning framework for the {Physical AI Spatial Intelligence Warehouse} dataset introduced in the Track 3 2025 {AI City Challenge}. Our approach enhances spatial comprehension by embedding mask dimensions in the form of bounding box coordinates directly into the input prompts, enabling the model to reason over object geometry and layout. We fine-tune the framework across four question categories namely: Distance Estimation, {Object Counting}, {Multi-choice Grounding}, and {Spatial Relation Inference} using task-specific supervision. To further improve consistency with the evaluation system, normalized answers are appended to the GPT response within the training set. Our comprehensive pipeline achieves a final score of \textbf{73.0606}, placing \textbf{4\textsuperscript{th}} overall on the public leaderboard. These results demonstrate the effectiveness of structured prompt enrichment and targeted optimization in advancing spatial reasoning for real-world industrial environments.
\end{abstract}    
\section{Introduction}
\label{sec:intro}

Spatial reasoning is a fundamental component of intelligent perception, enabling systems to interpret how objects relate within a 3D environment. In industrial settings such as warehouses, this capability is critical for tasks such as navigation, inventory management, and safety monitoring \cite{magee1989spatial}. These tasks rely on understanding object layouts, sizes, and relative distances to ensure safe and efficient operation. However, warehouse environments present additional challenges due to their cluttered and dynamic nature, with irregular structures, diverse object types, and frequent occlusions \cite{everett1995real, kim2023multi}. To operate effectively in such conditions, systems must capture both fine-grained visual details and the broader spatial organization of the scene. This need extends beyond recognition and requires methods that combine object detection with spatial understanding. Although computer vision has made substantial progress in detection \cite{asamoah2025saam, kyem2024advancing, aboah2023real} and segmentation \cite{AGYEIKYEM2025141583, wang2024gazesam, aboah2023deepsegmenter, AGYEIKYEM2026106591}, most existing approaches are tailored to isolated tasks and simplified environments. Their emphasis on local appearance limits their ability to model global spatial context, particularly in complex, real-world warehouse scenarios \cite{yoneyama2021integrating}.

\begin{figure}
    \centering
    \includegraphics[width=1\linewidth]{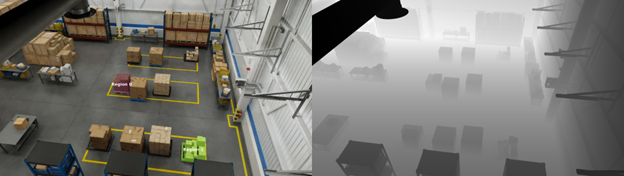}
    \begingroup
    \justifying
    \fontsize{7}{8.5}\selectfont

\noindent\textbf{Original Prompt:} 
    \textcolor{red}{'Is the pallet $<$mask$>$ to the left or right of the pallet $<$mask$>$?'} 

    \vspace{0.3em}
\noindent\textbf{Modified Prompt:} 
    \textcolor{blue}{Given all bounding box sizes are in the form x1y1x2y2, Is the pallet Region 0 within bounding box (139.2, 160.0, 160.6, 205.8) to the left or right of the pallet Region 1 within bounding box (222.8, 296.5, 253.4, 353.7)?'}  

    \vspace{0.3em}
\noindent\textbf{Original GPT answer:} 
    \textcolor{orange}{'The pallet [Region 0] is situated on the right of the pallet [Region 1].'} 

    \vspace{0.3em}
\noindent\textbf{Modified GPT answer:} 
    \textcolor{cyan}{'The pallet [Region 0] is situated on the right of the pallet [Region 1]. In short the normalized answer is right.'} 
    
    \caption{Example of spatial prompt transformation. The original prompt (top) uses natural language placeholders. The modified prompt encodes explicit bounding box coordinates, and the GPT-style answer is reformatted with a normalized response for consistent evaluation.}
    \label{fig:example}
    \endgroup
\end{figure}

Recent advances in Vision-Language Models (VLMs) have opened new avenues for spatial understanding by enabling joint reasoning over visual and textual inputs. Models such as BLIP-2~\cite{li2023blip} and InstructBLIP~\cite{dai2023instructblipgeneralpurposevisionlanguagemodels} support tasks like VQA~\cite{antol2015vqa}, image captioning~\cite{kyem2024pavecap}, and instruction following, but most rely on 2D imagery and lack explicit geometric grounding. This limitation restricts their effectiveness in tasks requiring spatial localization, physical comparison, or multi-object reasoning~\cite{cheng2024spatialrgpt, gu2024conceptgraphs}. While benchmarks like CLEVR~\cite{johnson2017clevr} and GQA~\cite{hudson2019gqa} test compositional reasoning in synthetic scenes \cite{10807178}, they fall short in capturing the structural complexity of real-world layouts. To address this, recent efforts have incorporated monocular depth estimation~\cite{bhat2021adabins, oquab2023dinov2} into VLM pipelines to provide geometric cues alongside semantic features~\cite{chen2024spatialvlm}. Though this integration improves performance in structured indoor settings, its effectiveness declines in industrial contexts like warehouses, where dense clutter, occlusion, and varied object scales challenge the spatial grounding capabilities of existing models~\cite{kim2023multi}.

To advance spatial reasoning in such environments, we build upon SpatialBot \cite{cai2024spatialbot}, a recent framework that integrates depth-aware encoding into vision-language models. We extend its functionality to better handle warehouse environments, which require fine-grained understanding of object arrangements, occlusions, and multi-object relationships. To improve the model's spatial comprehension, we introduce prompt-level enhancements that encode region masks as bounding box coordinates. As shown in Figure~\ref{fig:example}, our modified prompts replace vague descriptions with structured spatial cues, and we append the normalized answer to GPT responses to ensure consistency. We also fine-tune the model on the Physical AI Spatial Intelligence Warehouse dataset \cite{huggingfaceNvidiaPhysicalAISpatialIntelligenceWarehouseDatasets}, which contains complex layouts, varied object categories, and spatial questions that require both physical measurement and relational reasoning. Our approach improves the model's ability to answer spatial queries grounded in real-world warehouse structure and offers a practical path for applying hybrid depth-enhanced vision-language systems in industrial applications. To this end, we make the following contributions:

\vspace{0.02cm}
\begin{enumerate}
    \item We present a spatial question answering framework tailored to large-scale 3D industrial environments, leveraging spatially-informed prompts and grounded visual cues.
    \item We propose a prompt augmentation method that embeds object-level geometric features, including bounding box coordinates and mask dimensions, to enhance spatial reasoning.
    \item We extend the functionality of the SpatialBot architecture by fine-tuning it on the Physical AI Spatial Intelligence Warehouse dataset, enabling robust performance across four spatial reasoning tasks specific to cluttered warehouse layouts.
    \item We implement an output normalization module to align predictions with evaluation protocols, improving accuracy on fine-grained spatial categories.
    \item Our solution achieves a score of \textbf{73.0606} on the public leaderboard, placing \textbf{4\textsuperscript{th}} in Track 3 of the AI City Challenge 2025.
\end{enumerate}

\section{Related Work}
\label{sec:review}
Spatial reasoning plays a central role in vision-language systems, particularly for tasks involving object relationships, depth perception, and physical layout understanding. Prior work has made progress in vision-language modeling and monocular depth estimation, but their integration for fine-grained spatial understanding remains limited. This section reviews key developments across vision-language models, depth prediction, and spatial reasoning to position our work in the broader research landscape.

\subsection{Vision Language Models(VLMs)}
The success of large language models (LLMs) in NLP sparked interest in extending them to vision tasks, aiming to build unified models capable of multimodal reasoning. Visual Language Models (VLMs) have significantly advanced AI by combining vision and language understanding through large-scale multimodal training \cite{dai2023instructblipgeneralpurposevisionlanguagemodels, li2023blip, sun2019videobert, li2024videomamba}. These models, which pair a pre-trained LLM with a vision encoder, have shown strong performance across tasks like recognition and reasoning. Closed-source VLMs like GPT-4 \cite{achiam2023gpt}, Claude \cite{enis2024llm}, Gemini \cite{team2023gemini} and open models like, Video-Llama \cite{zhang2023video}, LLaVA \cite{liu2023visual} demonstrate comparable capabilities, largely due to their training on massive public and proprietary datasets. To solve the challenge of complex reasoning, studies have explored multi-modal chain-of-thought (CoT) reasoning \cite{zhang2023multimodal, chen2023see,BlessingAIC25,OworAIC25}, inspired by human problem-solving, where step-by-step rationales improve model performance. This includes using rich captions or multi-modal explanations for tasks like code generation \cite{li2025structured}, math \cite{fung2023chain}, and Question and answering \cite{huang2024coq}. Visual instruction tuning in VLMS, has also led to progress in perception \cite{michaelis2019benchmarking}, reasoning \cite{liu2024mmbench}, pixel-level grounding \cite{you2023ferret} and OCR \cite{li2024seed}. Despite these advances, most VLMs struggle with tasks requiring spatial localization and counting due to limited spatial grounding capabilities.
\vspace{-0.5em}
\subsection{Monocular depth estimation}
Monocular depth estimation has become a powerful tool for enhancing spatial understanding in vision systems. Early models relied on supervised learning with labeled datasets \cite{bhat2021adabins, silberman2012indoor} while later efforts adopted self-supervised strategies using stereo images \cite{godard2017unsupervised} or temporal consistency cues \cite{godard2017unsupervised}. Lately, large-scale pre-trained vision models have been fine-tuned for depth estimation using self-supervised \cite{oquab2023dinov2} and generative objectives \cite{rombach2022high} achieving strong performance on extensive depth datasets \cite{danier2025depthcues}. Two major types of monocular depth estimation is: discriminative and generative. Discriminative models directly regress depth from RGB inputs, often focusing on metric accuracy in specific domains like driving or indoor scenes. Techniques such as ordinal regression \cite{fu2018deep} and adaptive binning \cite{bhat2021adabins} have been used to improve accuracy. To enhance generalization, some methods estimate relative depth with scale-invariant losses \cite{ranftl2020towards} or integrate camera parameters as auxiliary inputs \cite{piccinelli2024unidepth}. In contrast, generative approaches, including latent diffusion models, capture finer scene details and structure \cite{shao2025learning}. While monocular depth methods offer promising geometric cues, integrating them into VLMs for spatial reasoning in cluttered environments remains an open challenge.
\subsection{Spatial Reasoning in Vision-Language Models}

Spatial reasoning is a critical yet underdeveloped capability in VLMs. Many existing models are trained solely on 2D image-text pairs \cite{li2023blip, chen2023pali}, which lack the depth information necessary for understanding geometric relationships and physical interactions in real-world scenes. This limitation is particularly evident in tasks requiring spatial localization or manipulation, such as those found in robotics. To address this gap, several approaches have emerged that augment VLMs to enhance their spatial understanding capabilities. For instance, SpatialVLM \cite{chen2024spatialvlm} and SpatialRGPT \cite{cheng2024spatialrgpt} enhance performance on spatial tasks by fine-tuning with curated datasets containing spatial questions and answers. However, these models primarily leverage linguistic input to guide spatial predictions, rather than learning spatial cues directly from visual signals. As a result, they often fall short when precise visual-grounded reasoning is required. Efforts to integrate spatial understanding into Large Language Models (LLMs) using 3D scene reconstructions or dense semantic features \cite{hong20233d} show promise, but they are often resource-intensive and struggle with modality alignment between vision and language. Alternatives like ConceptGraph \cite{gu2024conceptgraphs} attempt to bypass explicit 3D modeling by using structured scene graphs, yet studies show LLMs are not well-suited to reason over coordinate data embedded in text \cite{majumdar2024openeqa}. Monocular depth estimation has shown strong performance in estimating depth across diverse scenarios. SpatialBot \cite{cai2024spatialbot} enhances the spatial understanding of vision-language models (VLMs) by incorporating monocular depth estimation-generated depth into RGB inputs, addressing the challenge of inferring spatial context from a single image. In this work, we adopt SpatialBot \cite{cai2024spatialbot} for its demonstrated superiority in spatial intelligence.


\section{Methodology}

Our approach is designed to enhance spatial reasoning in complex 3D warehouse environments. The proposed system is built upon a vision-language model (VLM) that incorporates depth-aware encoding, segmentation-informed prompt augmentation, and instruction-based fine-tuning as shown in Figure \ref{fig:architecture}. This section outlines the core components of our methodology, including model architecture, training configuration, prompt processing, and answer normalization strategies.

\subsection{Model Architecture}
\begin{figure*}
    \centering
    \includegraphics[width=0.7\linewidth]{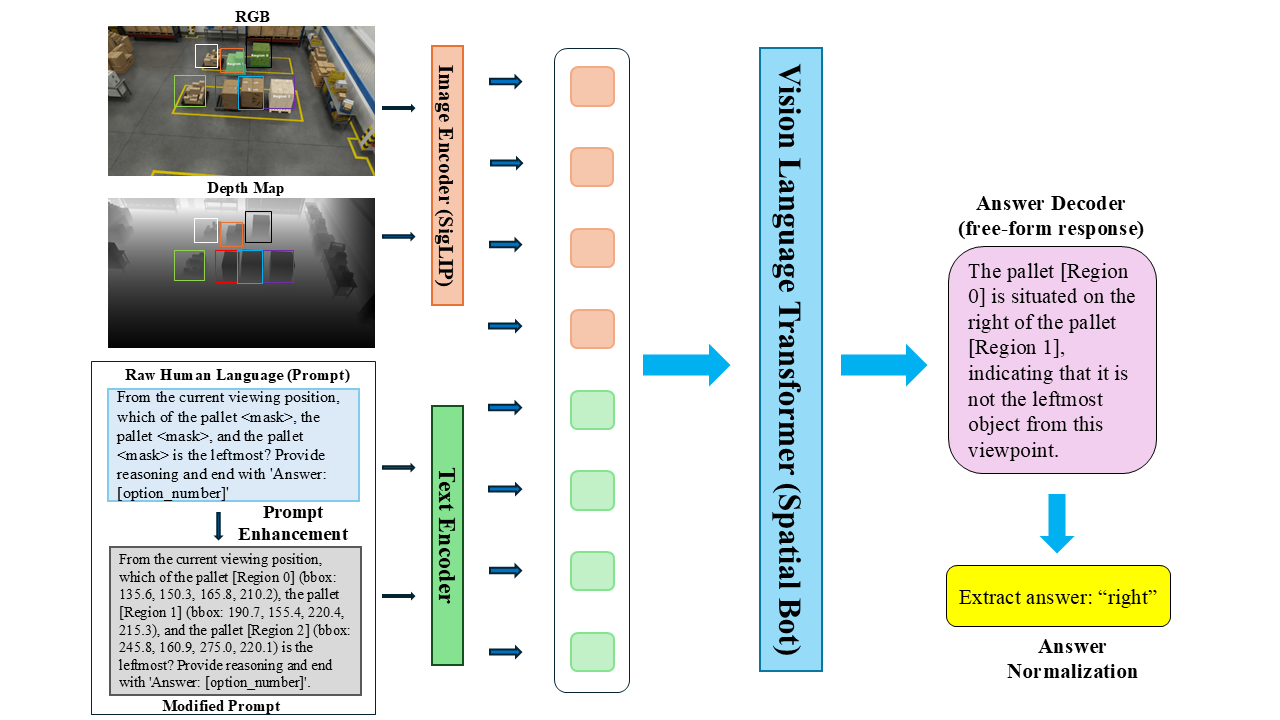}
    \caption{  Overview of our spatial reasoning architecture. The system processes RGB and depth images through a shared image encoder (SigLIP), while textual prompts are normalized and encoded separately. A vision-language transformer fuses the modalities to generate free-form responses. An answer normalization module extracts concise outputs. Spatial grounding is enabled by injecting bounding box coordinates and region identifiers into the prompts.}
    \label{fig:architecture}
\end{figure*}

We adopt SpatialBot~\cite{cai2024spatialbot}, a vision-language model (VLM) developed for spatial reasoning in cluttered indoor environments. The architecture (see Figure \ref{fig:architecture}) integrates an image encoder and a text encoder, which are jointly optimized with a lightweight language model. The image encoder takes both RGB and depth inputs, with depth images encoded into a three-channel \texttt{uint8} format. This representation helps the model capture fine-scale as well as wide-range spatial details. All input images are resized to $384\times384$ to meet the requirements of the pretrained encoders. In the original SpatialBot framework, several LLM backbones were evaluated, including Phi-2 (3B)~\cite{javaheripi2023phi}, Qwen-1.5 (4B)~\cite{bai2023qwen}, and LLaMA-3 (8B)~\cite{grattafiori2024llama}. Among these, Phi-2 was selected due to its strong balance between performance and model size, and we adopt the same configuration in our implementation.


\subsection{Prompt Enhancement}

The model is trained on instruction-formatted question-answer pairs derived from the Physical AI Warehouse dataset \cite{huggingfaceNvidiaPhysicalAISpatialIntelligenceWarehouseDatasets}. Each sample includes a spatial query and the corresponding response, designed to span multiple spatial reasoning tasks such as object counting, distance estimation, and directional inference. To enhance spatial grounding, we inject mask-derived dimensions in the form of bounding box information into the input prompts. Bounding box sizes are appended in the form of \texttt{x1, y1, x2, y2} for each relevant object, allowing the model to reason about relative positions. Furthermore, each bounding box is assigned a unique ID based on it's rank within a list of segmentation masks (e.g., “Region 0” for the first mask). This layout encoding provides the model with geometric context for each object pair in the scene.

\subsection{Answer Normalization}
During training, GPT-generated answers often follow a descriptive free-form format. However, evaluation requires a concise and normalized answer (e.g., “left”, “3”). To address this mismatch, we append a templated suffix—“In short, the normalized answer is \texttt{[label]}”—to the end of every training response. This ensures the model consistently embeds the required answer format during inference. An example transformation includes appending the string to both the question and answer before tokenization, preserving coherence between prompt and response during instruction tuning. The logic underlying this normalization strategy is formally outlined in Algorithm~\ref{alg:normalization}.

\begin{algorithm}[h!]
\caption{Answer Normalization for Spatial Reasoning}
\label{alg:normalization}
\scriptsize
\KwIn{$A$: Free-form GPT answer}  
\KwOut{$A_{\text{norm}}$: Normalized short answer (e.g., \texttt{left}, \texttt{right}, \texttt{9.81 meters})}

\If{``In short, the normalized answer is'' $\in A$}{
    Extract substring after ``In short, the normalized answer is''\;
    Remove punctuation and convert to lowercase\;
    $A_{\text{norm}} \leftarrow$ cleaned substring\;
}
\ElseIf{$A$ contains spatial cue (\texttt{left}, \texttt{right}, \texttt{meters})}{
    Match most probable directional or numeric phrase\;
    $A_{\text{norm}} \leftarrow$ extracted cue\;
}
\Else{
    Flag $A$ for manual review or fallback heuristics\;
}
\Return{$A_{\text{norm}}$}\;
\end{algorithm}


\vspace{-1.5em}
\section{Experiment}

\subsection{Dataset}
This study uses the Physical AI Spatial Intelligence Warehouse dataset \cite{huggingfaceNvidiaPhysicalAISpatialIntelligenceWarehouseDatasets}, introduced by NVIDIA \cite{naphade2017nvidia} to support spatial reasoning in warehouse-scale 3D environments. The dataset was created using the NVIDIA Omniverse platform \cite{hummel2019leveraging} and consists of approximately 95,000 RGB-D image pairs paired with over 499,000 question-answer (QA) pairs for training, 19,000 QA pairs for testing, and 1,900 for validation. Each data point includes an RGB image, a depth map, an object mask, and a natural language QA pair with a normalized single-word answer. The questions are designed to test spatial understanding across four categories: spatial relationships (e.g., left/right), multi-choice target identification, distance estimation between objects, and object counting. Annotations are automatically generated using rule-based templates and refined with large language models to produce more natural language responses. All object-level labels and region masks are synthesized using NVIDIA IsaacSim \cite{nvidiaIsaac}.

\subsection{Evaluation Metrics}
The primary metric for Track 3 is the \textbf{Weighted Average Success Rate (WASR)}, which measures the overall percentage of correctly answered questions across all categories. A prediction is considered correct (success = 1) if it meets the required criterion per task; otherwise, success = 0.

\noindent\textbf{Weighted Average Success Rate} is computed as:
\begin{equation}
\text{WASR} = \frac{1}{N} \sum_{i=1}^{N} \mathbb{1} \left[ \text{Prediction}_i \in \text{Valid}_i \right]
\end{equation}

\noindent where $N$ is the total number of questions, and $\mathbb{1}[\cdot]$ is an indicator function that equals 1 if the prediction is valid under the task-specific evaluation rule.

\vspace{0.5em}
\noindent\textbf{Distance and Count} questions are evaluated using \textbf{Acc@10}, where a prediction is successful if it lies within the top 10 closest answers (based on confidence) and satisfies:
\begin{equation}
\left| \frac{\text{Prediction} - \text{GT}}{\text{GT}} \right| \leq 0.10
\end{equation}

\vspace{0.5em}
\noindent\textbf{Multiple-Choice} and \textbf{Spatial Relation} tasks are evaluated using standard accuracy:
\begin{equation}
\text{Accuracy} = \frac{\text{Number of Correct Predictions}}{\text{Total Predictions}}
\end{equation}

\vspace{0.5em}
\noindent\textbf{Relative Error} is also reported for Distance and Count questions to support fine-grained analysis:
\begin{equation}
\text{Relative Error} = \frac{|\text{Prediction} - \text{GT}|}{\text{GT}} \times 100\%
\end{equation}

\vspace{0.5em}
\noindent\textbf{Answer Normalization} is applied to reduce variability in responses. Predictions are mapped to a canonical format that accounts for case, digits, and unit consistency. For example, ``Four'', ``4'', and ``4.0'' are all interpreted as equivalent.

\subsection{Training Configuration}
\label{sec:implementation}

To balance computational constraints with model performance, the \textit{SpatialBot} model was fine-tuned using a subset of 100{,}000 prompts randomly sampled from the full dataset of approximately 500,000 instances. Fine-tuning was conducted over 12500 iterations using the AdamW optimizer \cite{pytorchAdamWx2014} with a learning rate of $2 \times 10^{-4}$, weight decay of 0.01, and a batch size of 8. To reduce memory usage and training time, LoRA fine-tuning \cite{li2023loftq} was applied with a rank of 128 and an alpha scaling factor of 256. When run on 2 NVIDIA A40 GPUs (48GB each), the training time per epoch was approximately 127 hours.

\section{Results and Discussion}
\label{sec:results}

\subsection{Quantitative Evaluation}
We evaluate two vision-language models on the {Physical AI Spatial Intelligence Warehouse} dataset: {Qwen-VL-2.5 \cite{yang2025qwen2}}, a general-purpose visual instruction model, and {SpatialBot \cite{cai2024spatialbot}}, a depth-enhanced model optimized for spatial reasoning. The evaluation covers all four official task categories: {Object Counting}, {Distance Estimation}, {Left-Right Reasoning}, and {Multi-Choice Grounding}, along with aggregated scores for {Quantitative}, {Qualitative}, and the final benchmark metric {S1}.

Table~\ref{tab:comparison} summarizes the full test set results. {SpatialBot} achieves the highest overall performance with an S1 score of \textbf{73.06}, compared to {31.92} from Qwen-VL-2.5. SpatialBot records strong accuracy across tasks, including {Left-Right Reasoning (99.7000)}, {Qualitative (83.9703)}, and {Quantitative (63.2565)} categories. It also achieves low error rates for {Count RMSE (0.2320)} and {Distance RMSE (1.3380)}. In contrast, Qwen-VL-2.5 struggles with spatial generalization. Despite its broad instruction-following capabilities, its performance remains limited across all categories especially distance measurement. The authors hypothesize that this is due to the base model’s initial approach to depth estimation, which utilized depth points rather than meters, as used in the training set. This discrepancy, an inherent attribute of the base model, would require additional training to overcome.

\begin{table}[h]
\centering
\caption{Performance comparison on the Physical AI Warehouse dataset.}
\label{tab:comparison}
\renewcommand{\arraystretch}{1.1}
\setlength{\tabcolsep}{2.8pt}  
\scriptsize  
\begin{tabular}{lccccccccc}
\toprule
\textbf{Model} & \textbf{Cnt} & \textbf{RMSE} & \textbf{Dist} & \textbf{D-RMSE} & \textbf{LR} & \textbf{MCQ} & \textbf{Quant} & \textbf{Qual} & \textbf{S1} \\
\midrule
Qwen-VL  & 37.96 & 0.763 & 13.30 & 3.643 & 62.02 & 14.42 & 25.92 & 39.28 & 31.92 \\
SpatialBot   & 78.81 & 0.232 & 46.95 & 1.338 & 99.70 & 66.78 & 63.26 & 83.97 & 73.06 \\
\bottomrule
\end{tabular}
\end{table}



\subsection{Qualitative Evaluation}
\begin{figure}
    \centering
    \includegraphics[width=1\linewidth]{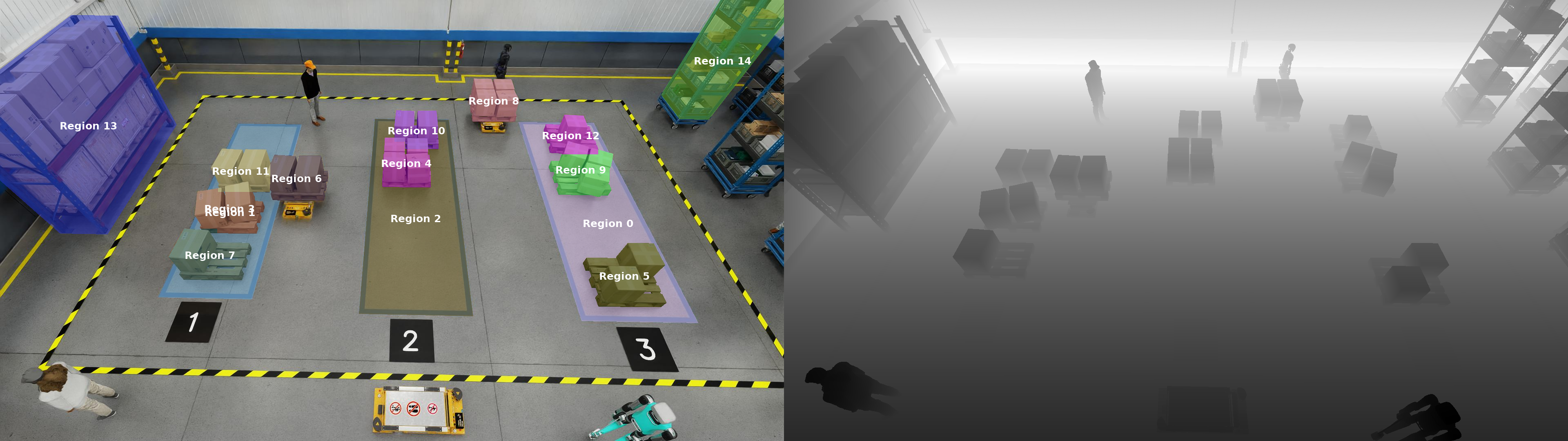}
    \begingroup
    \justifying
    \fontsize{7}{8.5}\selectfont

    \vspace{0.3em}
\noindent\textbf{Modified Prompt:} 
    \textcolor{red}{Given all bounding box sizes are in the form x1y1x2y2. Using the buffer masks Region 0 within bounding box (451.5, 59.8, 607.6, 158.0) Region 1 within bounding box (137.9, 60.6, 262.4, 146.4) Region 2 within bounding box (312.5, 58.2, 411.7, 154.8) and pallet masks Region 3 within bounding box (169.6, 89.0, 230.0, 114.2) Region 4 within bounding box (332.8, 67.4, 374.75, 91.6) Region 5 within bounding box (507.37, 119.0, 579.9, 150.2) Region 6 within bounding box (231.1, 75.8, 285.1, 98.0) Region 7 within bounding box (146.8, 112.0, 218.6, 137.0) Region 8 within bounding box (408.1, 38.4, 451.5, 59.4) Region 9 within bounding box (477.8, 69.0, 533.3, 96.2) Region 10 within bounding box (343.4, 54.0, 381.5, 72.8) Region 11 within bounding box (183.82, 72.8, 235.0, 93.8) Region 12 within bounding box (473.2, 56.2, 520.5, 75.4), how many pallets are situated in the buffer region closest to the shelf on the right among Region 13 within bounding box (0.0, 7.4, 153.6, 114.6) Region 14 within bounding box (575.6, 0.0, 682.3, 58.4)? Please give the normalized answer only.
'}  

    \vspace{0.3em}
\noindent\textbf{Ground Truth:} 
    \textcolor{orange}{The shelf [Region 14] is the shelf on the right. The buffer region [Region 0] is the closest to the shelf [Region 14]. I see pallets [Region 5] [Region 9] [Region 12] in the buffer region [Region 0]. Hence, in buffer area [Region 0], there are exactly three pallets.} 

    \vspace{0.3em}
\noindent\textbf{Model Prediction:} 
    \textcolor{blue}{The buffer region [Region 1] is the closest to the shelf [Region 14]. There are pallets [Region 3] [Region 8] [Region 10] in the buffer region [Region 1]. Hence, in buffer area [Region 1], there are exactly three pallets. In short the normalized answer is “3”.} 

    
    \caption{Qualitative example illustrating the model's ability to count pallets within the buffer region closest to the rightmost shelf. The model correctly identifies Region 14 as the shelf, Region 1 as the closest buffer zone, and detects three relevant pallet regions within the specified area.}
    \label{fig:prompt}
    \endgroup
\end{figure}

\begin{figure}
    \centering
    \includegraphics[width=1\linewidth]{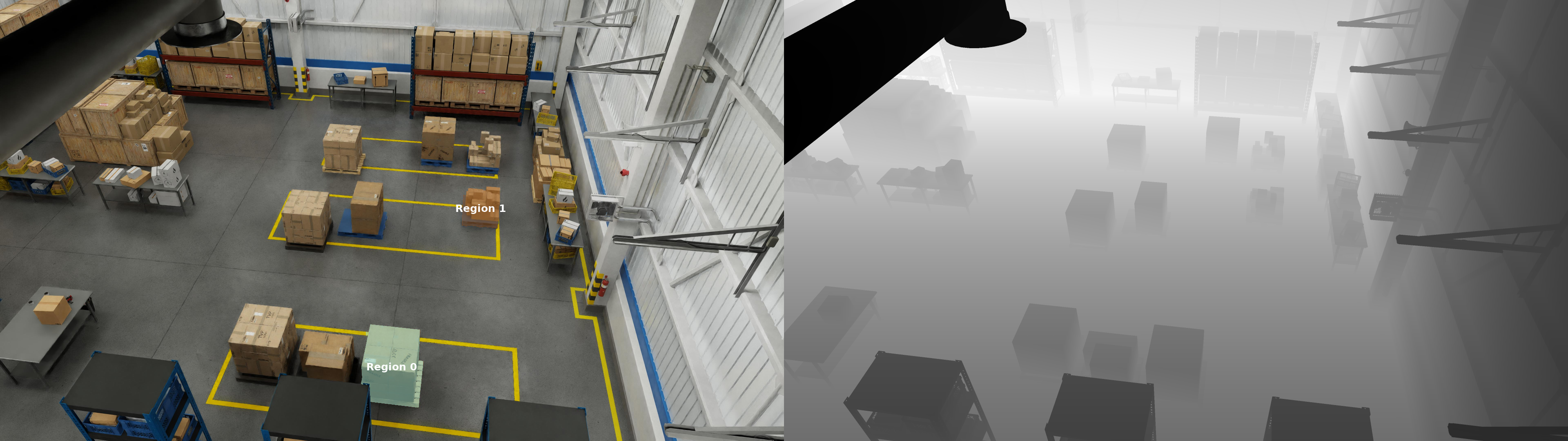}
    \begingroup
    \justifying
    \fontsize{7}{8.5}\selectfont

    \vspace{0.3em}
\noindent\textbf{Modified Prompt:} 
    \textcolor{red}{Given all bounding box sizes are in the form x1y1x2y2Can you determine if the pallet Region 0 within bounding box (314.31111111111113, 158.8, 368.0, 199.4) is to the right of the pallet Region 1 within bounding box (402.1333333333333, 91.4, 434.84444444444443, 111.6) based on the current viewing angle? Please give the normalized answer only.}  

    \vspace{0.3em}
\noindent\textbf{Ground Truth:} 
    \textcolor{orange}{Looking from this angle, the pallet [Region 0] is to the left of the pallet [Region 1].} 

    \vspace{0.3em}
\noindent\textbf{Model Prediction:} 
    \textcolor{blue}{From the image's perspective, the pallet [Region 0] is on the left of the pallet [Region 1]. In short the normalized answer is “left”.} 

    
    \caption{Qualitative example demonstrating the model’s capability in pairwise spatial comparison. The model accurately infers that Region 0 lies to the left of Region 1 from the given viewpoint and bounding box inputs.}
    \label{fig:prompt2}
    \endgroup
\end{figure}

To illustrate the model’s spatial reasoning ability, we present two examples in Figure \ref{fig:prompt} and Figure \ref{fig:prompt2} that cover counting and comparison tasks. In Figure \ref{fig:prompt}, the model is asked to count how many pallets are in the buffer region nearest to the shelf on the right. It correctly identifies Region 14 as the shelf and Region 1 as the closest buffer. Within that region, it detects three pallets i.e Region 3, Region 8, and Region 10. The predicted answer, “3”, aligns with the ground truth, showing the model’s ability to reason over multiple regions using geometric and positional cues.

Figure \ref{fig:prompt2} focuses on a pairwise comparison task. The model is asked to determine which of two pallets is on the left from the current viewpoint. Based on the bounding box positions, it correctly identifies that Region 0 is to the left of Region 1. The predicted answer, “left”, matches the ground truth and confirms the model’s capacity to reason about spatial layout with respect to viewpoint. These examples demonstrate that the model can perform both fine-grained comparisons and broader spatial reasoning involving multiple objects in complex scenes.

\subsection{Ablation Study}
Table~\ref{tab:abla} presents an ablation study evaluating the impact of bounding box grounding on overall model performance. SpatialBot\_v1, which does not incorporate bounding box grounding, achieves an S1 score of 47.69. In contrast, SpatialBot\_v2 integrates explicit grounding and yields a substantial performance gain, achieving an S1 score of 73.06. This significant improvement highlights the critical role of spatial grounding in aligning visual object regions with natural language queries, demonstrating that incorporating bounding box grounding leads to more accurate and context-aware reasoning in physical AI tasks.

\begin{table}[h]
\centering
\caption{Ablation Study.}
\label{tab:abla}
\renewcommand{\arraystretch}{1.1}
\setlength{\tabcolsep}{5pt}
\scriptsize
\begin{tabular}{lc|cc}
\toprule
\textbf{Model} & \textbf{Bounding Box Grounding} & \textbf{S1 Score} \\
\midrule
SpatialBot\_v1  & X & 47.69 \\
SpatialBot\_v2   & \checkmark & 73.06 \\
\bottomrule
\end{tabular}
\end{table}

\subsection{Experimental Test Dataset}
Table~\ref{tab:warehouse_leaderboard} presents the final results from the 2025 AI City Challenge Track 3, evaluated using the S1 Score across the full test set. Our method achieved an \textbf{S1 Score of 73.0606}, placing \textbf{4\textsuperscript{th} among all teams}. This result highlights the strength of our spatially guided vision-language approach and its ability to handle complex reasoning tasks in cluttered, real-world warehouse scenes. Our competitive placement reinforces the value of depth-enhanced prompt design for advancing spatial understanding in logistics-scale environments.

\begin{table}[h]
\centering
\caption{Top 5 Teams in the Warehouse Spatial Intelligence Track}
\begin{tabular}{|c|c|l|c|}
\hline
\textbf{Rank} & \textbf{Team Name} & \textbf{Score} \\
\hline
1 & UWIPL\_ETRI       & 95.8638 \\
2 & HCMUT.VNU         & 91.9735 \\
3 & Embia             & 90.6772 \\
4 & \textbf{MIZSU (Ours)}     & \textbf{73.0606} \\
5 & HCMUS\_HTH        & 66.8861 \\
\hline
\end{tabular}
\label{tab:warehouse_leaderboard}
\end{table}


\subsection{Limitations}
Within the constraints of this challenge, the authors note two key obstacles encountered. Firstly, model training was time-consuming, despite the adoption of LoRA, the use of pretrained weights, and access to relatively large GPU resources. This constraint effectively “prices out” fine-tuning and adoption of these models by those with limited computational resources. Moreover, it restricts experimentation with different large language model (LLM) architectures.
Secondly, participants highlight the importance of inference time under competition constraints. For instance, running inference on the entire test dataset (19,000 images) using a 16 GB GPU and 64 GB RAM took approximately 12 hours. This further limits the ability to experiment with different approaches.

\section{Conclusion}
\label{sec:conclusion}


This work introduces a spatial reasoning approach designed to handle the visual complexity of real-world warehouse environments. The method incorporates monocular depth maps and spatially enriched prompts containing bounding box coordinates, enabling the vision-language model to better capture object arrangements and spatial relationships. The system is further refined through task-specific fine-tuning on a diverse warehouse benchmark featuring physical measurements and multi-object comparisons.

Evaluated on the 2025 AI City Challenge Track 3, the approach achieved a final score of \textbf{73.0606}, securing \textbf{4th place on the public leaderboard}. These results demonstrate the effectiveness of integrating geometric priors and prompt-level enhancements for fine-grained spatial understanding. The proposed solution offers a practical direction for applying depth-augmented vision-language reasoning in cluttered industrial settings.

{
    \small
    \bibliographystyle{ieeenat_fullname}
    \bibliography{main}
}


\end{document}